\documentclass[11pt,a4paper]{article}
\usepackage[hyperref]{acl2019}
\usepackage{times}
\usepackage{latexsym}

\usepackage{url}
\usepackage[]{algorithm2e}

\usepackage{amsmath, amsthm, amssymb}    
\usepackage{amsthm}
\usepackage{graphicx}
\usepackage{amsmath}
\usepackage{amsfonts}
\usepackage{graphicx}
\usepackage{verbatim}
\usepackage{booktabs}
\usepackage{subfig}
\usepackage{amsthm}
\usepackage{amssymb}
\usepackage{qtree}
\usepackage{graphicx}
\usepackage{caption}
\usepackage{multirow}
\usepackage{mathrsfs}
\usepackage{tablefootnote}
\usepackage{arydshln}
\usepackage{lscape}
\usepackage{afterpage}
\usepackage{todonotes}
\usepackage{arydshln}
\usepackage{color}
\usepackage[linewidth=1pt]{mdframed}
\usepackage{enumitem}
\usepackage{textcomp}

\usepackage{url}
\usepackage{caption}
\usepackage{xcolor}
\usepackage{calc}

\usepackage{pgfplots} 
  
\aclfinalcopy 

\newcommand\numberthis{\addtocounter{equation}{1}\tag{\theequation}}

\title{Boosting Entity Linking Performance by Leveraging \\ Unlabeled Documents}

\author{Phong Le$^1$ \and Ivan Titov$^{1,2}$ \\
  $^{1}$University of Edinburgh   $\;^{2}$University of Amsterdam \\
  {\tt lephong.xyz@gmail.com $\;\;$ ititov@inf.ed.ac.uk}
  \\}

\date{}

\begin{document}
\maketitle
\begin{abstract}
Modern entity linking systems rely on large collections of documents specifically annotated for the task (e.g., AIDA CoNLL). In contrast, we propose an approach which exploits only naturally occurring information: unlabeled documents and Wikipedia. Our approach consists of two stages. First, we construct a high recall list of candidate entities for each mention in an unlabeled document. Second, we use the candidate lists as weak supervision to constrain our document-level entity linking model. The model treats entities as latent variables and, when estimated on a collection of unlabelled texts, learns to choose entities
relying both on local context of each mention and on coherence with other entities in the document.
The resulting approach rivals fully-supervised state-of-the-art systems on standard test sets.  
It also approaches their performance in the very challenging setting: 
when tested on a test set sampled from the data used to estimate the supervised systems.
By comparing to Wikipedia-only training of our model, we demonstrate
that modeling unlabeled documents is beneficial.

\end{abstract}


\section{Introduction}
\label{section intro}

\begin{figure}[t!]
\centering
\includegraphics[width=0.45\textwidth]{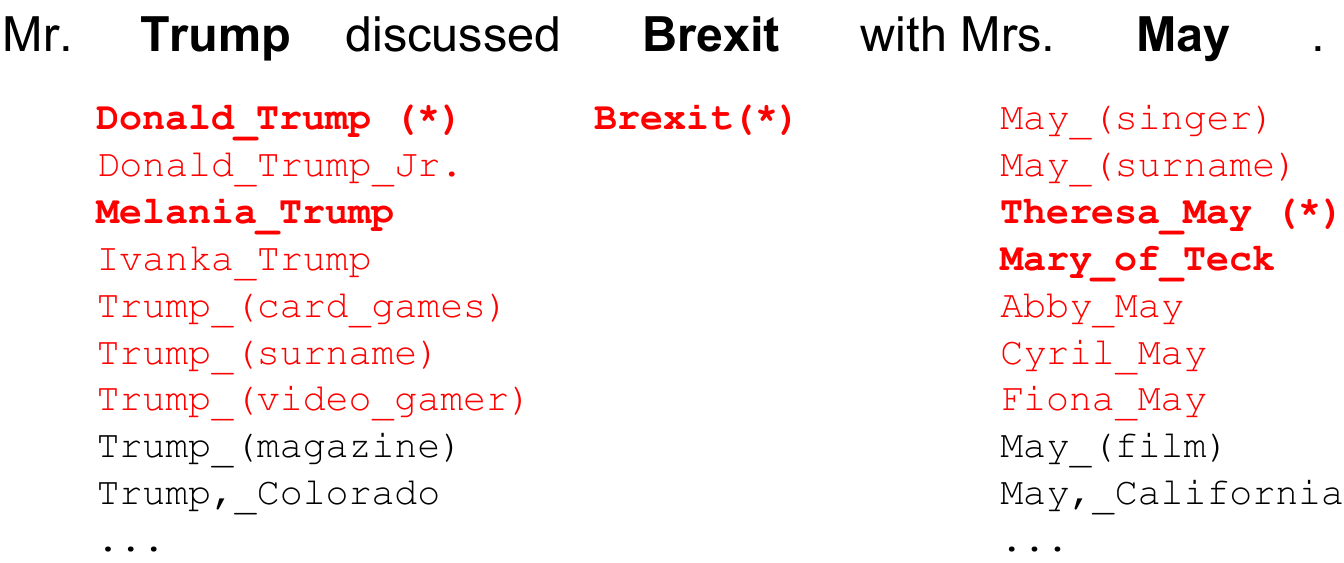}
\caption{A sentence with candidate entities for mentions. 
The correct entities are marked with (*). 
We \textit{automatically} extract likely candidates (red bold) 
and likely negative examples (non-bold red).
These are used to train our weakly-supervised model.} 
\label{fig intro}
\end{figure}

Named entity linking is the task of linking a mention to the corresponding 
entity in a knowledge base (e.g., Wikipedia). For instance, in Figure~\ref{fig intro}
we link mention ``Trump'' to Wikipedia entity \texttt{Donald\_Trump}. 
Entity linking enables aggregation of information across multiple mentions of the same entity which is crucial in many natural language processing applications
such as question answering~\cite{hoffmann-EtAl:2011:ACL-HLT2011,welbl2018constructing},
information extraction~\cite{hoffmann-EtAl:2011:ACL-HLT2011} or multi-document summarization \cite{nenkova2008entity}.



While traditionally entity linkers relied mostly on Wikipedia and heuristics~\cite{milne2008learning,P11-1138,cheng-roth:2013:EMNLP}, the recent generation of methods \cite{P16-1059,guorobust,K16-1025,D17-1276,2018arXiv180410637L} approached the task as supervised learning on
a collection of documents specifically annotated for the entity linking problem
(e.g.,  relying on AIDA CoNLL \cite{hoffart-EtAl:2011:EMNLP}). While they substantially outperform the traditional methods, such human-annotated resources are scarce (e.g., available mostly for English) and expensive to create.  
 Moreover, the resulting models end up being domain-specific:
their performance drops substantially when they are used in a new domain.\footnote{The best reported in-domain scores are 93.1\% F1 \cite{2018arXiv180410637L}, whereas the best previous  out-of-domain score is only 85.7\% F1 ~\cite{guorobust} (an average over 5 standard out-of-domain test sets, Table ~\ref{tab:results}).}
We will refer to these systems as {\it fully-supervised}.


Our goal is to show that an accurate entity linker can be created relying solely on naturally occurring data. Specifically, our approach relies only on Wikipedia and a collection of unlabeled texts. Though links in Wikipedia have been created by humans,  no extra annotation is necessary  to build our linker. Wikipedia is also available in many languages and covers many domains. Though Wikipedia information is often used within entity linking pipelines, previous systems relying on Wikipedia are substantially less accurate than modern fully-supervised systems 
 (e.g., Cheng and Roth~\shortcite{cheng-roth:2013:EMNLP}, Ratinov at al.~\shortcite{P11-1138}). This is also true
of the only other method which, like ours, uses a combination of Wikipedia data  and unlabeled texts~\cite{TACL637}. We will refer to approaches using this form of supervision, including our approach, as {\it Wikipedia-based linkers}.

Wikipedia articles have a specific rigid structure~\cite{chen2009content}, often dictated by the corresponding templates, and mentions in them are only linked once (when first mentioned). For these reasons,  Wikipedia pages were
not regarded as suitable for training document-level models \cite{P16-1059,D17-1276}, whereas state-of-the-art fully supervised  methods rely on document-level modeling.
We will show that, by exploiting unlabeled documents and estimating  document-level neural coherence models on these documents, we can bring Wikipedia-based linkers on par or, in certain cases, make them more accurate than fully-supervised linkers.

Our Wikipedia-based approach uses two stages: candidate generation and document-level disambiguation. First, we take an unlabeled document collection and  use link statistics in Wikipedia to construct a high recall list of candidates for each mention in each document. To create these lists, we use the Wikipedia link graph, restrict vertices to the ones
potentially appearing in the document (i.e. use the `vertex-induced subgraph' corresponding to the document) and perform message passing with a simple probabilistic model which does not have any trainable parameters.
After this step, for the example in Figure \ref{fig intro}, we would be left with \texttt{Theresa\_May} and a Queen of England  \texttt{Mary\_of\_Teck} as two potential candidates for mention ``May,'' whereas we would rule out many other possibilities (e.g., a former settlement in California). 
Second, we train a document-level statistical disambiguation model which treats entities as latent variables and uses the candidate lists as weak supervision. Intuitively, the disambiguation model is trained to score at least one assignment compatible with the candidate lists higher than all the assignments incompatible with the lists (e.g., one which links ``Trump'' to {\tt Ivanka\_Trump}). Though 
the constraints do not prevent linking ``May'' to the Queen in Figure \ref{fig intro}, given enough data, the model should rule out this assignment as not in fitting with other entities in the document (i.e. {\tt Donald\_Trump} and {\tt Brexit})
and/or not compatible with its local context  (i.e. ``Mrs.'').

We evaluate our model against previous methods on six standard test sets, covering multiple domains. Our model achieves the best results on four of these sets and in average. Interestingly, our system performs well on test data from AIDA CoNLL,
the dataset used to train fully-supervised systems, even though we have not used the annotations. 

Our approach also substantially outperforms both previous Wikipedia-based approaches and a version of our system which is simply trained to predict Wikipedia links. 
This result demonstrates that unlabeled data was genuinely beneficial. 
We perform ablations confirming that 
the disambiguation model benefits from capturing both coherence with other entities (e.g., \texttt{Theresa\_May} is more likely than \texttt{Mary\_of\_Teck} to appear in a document mentioning \texttt{Donald\_Trump}) and from exploiting local context of mentions (e.g., ``Mrs.'' can be used to address a prime minister but not a queen). This experiment confirms an intuition that global modeling of unlabeled documents is preferable to training local models to predict individual Wikipedia links.
Our contributions can be summarized as follows: 
\begin{itemize}
\item we show how Wikipedia and unlabeled data can be used to construct an accurate linker which rivals linkers constructed using expensive human supervision;
\item we introduce a novel constraint-driven approach to learning a document-level (`global') co-reference model without using any document-level annotation;
\item we provide evidence that fully-annotated documents
 may not be as beneficial as previously believed.
\end{itemize}

\section{Constraint-Driven Learning for Linking}
\label{section model}

\subsection{Setting}

We assume that for each mention $m_i$, we are provided with a set of candidates $E^+_i$. In subsequent section we will clarify how these candidates are produced.
For example, for $m_1=$``Trump'' in Figure~\ref{fig intro}, the set would be $E^+_{1}=\{Donald\_Trump,Melania\_Trump\}$. When learning our model we will assume that one entity candidate in this set is correct ($e^{*}_i$). 
Besides the `positive examples'  $E^+_i$, we assume that we are given a set of wrong entities $E^-_i$ (including, in our example, \texttt{Ivanka\_Trump} and \texttt{Donald\_Trump\_Jr}). 

In practice our candidate selection procedure 
is not perfect and  the correct entity $e^{*}_i$ will occasionally be missed from $E^+_i$ and even misplaced into $E^-_i$. This is different from the standard supervised setting where $E^{+}_i$ contains a single entity, and the annotation is not noisy. Moreover, unlike the supervised scenario, we do not aim to learn to mimic the teacher but rather want to improve on it relying on other learning signals (i.e. document context). 

Some mentions do not refer to any entity in a knowledge base and should, in principle, be left unlinked. In this work, we  link mentions whenever there are any candidates for linking them. More sophisticated ways of dealing with {\it NIL-linking} are left for future work.

\subsection{Model}
\label{sect:model}

Our goal is to not only model fit between an entity and its local context but also model  interactions between entities in a document (i.e. coherence between them).
As in previous global entity-linking models~\cite{P11-1138}, we can define the scoring function for $n$ entities $e_1$, \ldots, $e_n$ in a document $D$ as a conditional random field:
\begin{equation*}
g(e_1, \ldots, e_n | D) = \sum_{i=1}^{n}{ \phi(e_i | D)} + \sum_{j \neq i}{ \psi(e_i, e_j | D)},
\end{equation*}
where the first term scores how well an entity fits the context and the second one judges coherence.
Exact MAP (or max marginal) inference, needed both at training and testing time, is NP-hard~\cite{wainwright2008graphical}, and even approximate methods (e.g., loopy belief propagation, LBP) are relatively expensive and do not provide convergence guarantees. Instead, we score entities independently relying on the candidate lists: 
\begin{equation}
\label{equ model}
s(e_i | D) = \phi(e_i | D) + \sum_{j \neq i}{ \max_{e_j \in E^+_j} \psi(e_i, e_j | D)}.
\vspace{-1ex}
\end{equation}
Informally, we score $e_i$ based on its coherence with the `most compatible' candidate for each mention in the document. This scoring strategy is computationally efficient
and has been shown effective in the supervised setting by Globerson et al.~\shortcite{P16-1059}.
They refereed to this approach as a `star model', as it can be regarded as exact inference in a modified graphical model.\footnote{For each $e_i$,
you create its own graphical model: keep only edges connecting $e_i$ to all other entities; what you obtain is a star-shaped graph with $e_i$ at its center.}

%

\begin{figure}
	\centering
    \includegraphics[width=0.45\textwidth]{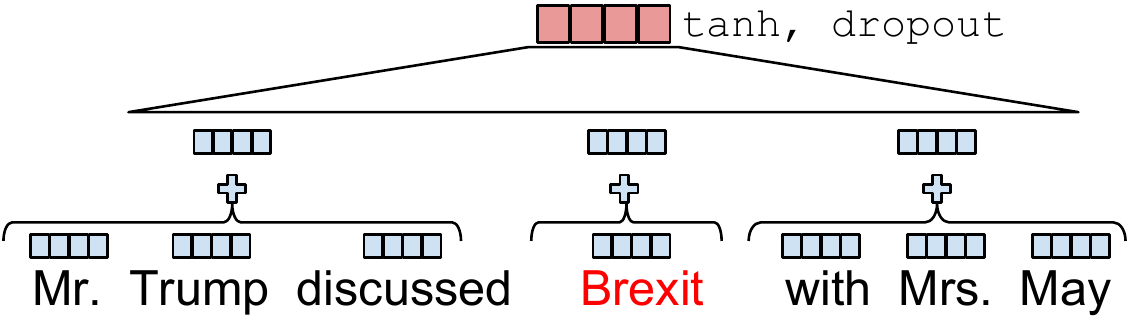}
    \caption{$\mathbf{h}(m_i, c_i)$ is a one-layer neural network, 
    with $\tanh$ activation and a layer of dropout on top.}
    \label{fig:multi-relation-a}
\end{figure}

We instantiate the general model for the above expression~(\ref{equ model}) 
in the following form:
\begin{equation}
\nonumber
s(e_i | D) = \phi(e_i | c_i, m_i) +  \sum_{j \neq i} \alpha_{ij} { \max_{e_j \in E^+_j} {
\xi(e_i, e_j)
}},
\end{equation}
where we use $m_i$ to denote an entity mention, $c_i$ is its context (a text window around the mention),  $\xi(e_i,e_j)$
is a pair-wise compatibility score and $\alpha_{ij}$ are attention weights, measuring relevance of an entity at position $j$ to predicting entity $e_i$ (i.e. $\sum_{j=1}^{n}{\alpha_{ij}} = 1$). The local score $\phi$ is identical to the one used 
in Ganea and Hofmann~\shortcite{D17-1276}. 
 As the pair-wise compatibility score we use
$\xi (e_i, e_j) = \mathbf{x}^T_{e_i} \mathbf{R} \mathbf{x}_{e_j}, $
where $\mathbf{x}_{e_i}$ and $\mathbf{x}_{e_j} \in \mathbb{R}^{d_e}$ are external entity embeddings, which are not fine-tuned in training. $\mathbf{R} \in \mathbb{R}^{d_e \times d_e}$ is a diagonal matrix. The attention is computed as
\begin{equation*}
	\alpha_{ij} \propto \exp\left\{\mathbf{h}(m_i,c_i)^T \mathbf{A} \mathbf{h}(m_j,c_j)/\sqrt{d_c}\right\} 
\end{equation*}
where the function $\mathbf{h}(m_i, c_i)$ mapping a mention and its context to $\mathbb{R}^{d_c}$ is given in Figure~\ref{fig:multi-relation-a}, $\mathbf{A} \in \mathbb{R}^{d_c \times d_c}$ is a diagonal matrix. A similar attention model was used in the supervised linkers of Le and Titov~\shortcite{2018arXiv180410637L} and Globerson et al.~\shortcite{P16-1059}.

Previous supervised methods such as Ganea and Hofmann~\shortcite{D17-1276} additionally exploited 
a simple extra feature
$p_{wiki}(e_i|m_i)$: the normalized frequency of mention $m_i$ being used as an anchor text for entity $e_i$ in Wikipedia articles and YAGO. We combine this score with the model score $s(e_i | D)$ using a one-layer neural network to yield  $\hat{s}(e_i | D)$.  At test time, we use our model to select entities from the candidate list.
As standard in reranking~\cite{collins2005discriminative}, we linearly combine $\hat{s}(e_i | D)$ with the score $s_c(e_i | D)$  from the candidate generator, defined below (Section \ref{subsection candidate score}).\footnote{We do not train the linear coefficient in an end-to-end fashion, as we do not want our model to over-rely on the candidate selection procedure at training time.}  
The hyper-parameters are chosen using a development set. Additional details are provided in the appendix.



\subsection{Training}
\label{subsection training}
As we do not know which candidate in $E^+_i$ is correct, 
we train the model to score at least one candidate in $E^+_i$ higher than any negative example from $E^-_{i}$. 
This approach is reminiscent of constraint-driven learning~\cite{chang2007guiding}, 
as well as of multi-instance learning methods common in 
relation extraction~\cite{riedel2010modeling,surdeanu2012multi}.
Specifically, we minimize 

\begin{align*}
L(\Theta) = \sum_D \sum_{m_i} \Big[&\delta + 
\max_{e^-_i \in E^-_i} \hat{s}(e^-_i|D) \\
& - \max_{e^+_i \in E^+_i} \hat{s}(e^+_i|D) \Big]_+ 
\end{align*}
where $\Theta$ is the set of model parameters, $\delta$ is a margin, 
and $[x]_+=\max\{0, x\}$.

\section{Producing Weak Supervision}
\label{section dl for el}


We rely primarily on Wikipedia to produce weak supervision.
We  start with a set of candidates for a mention $m$ containing all entities refereed to with anchor text $m$ in Wikipedia. We then filter this set in two steps.
The first step is the preprocessing technique of Ganea and Hofmann~\shortcite{D17-1276}. 
After this  step, the list has to remain fairly large in order to maintain high recall. Large lists are not effective as weak supervision as they do not sufficiently constraint the space of potential assignments to drive learning of the entity disambiguation model.
In order to further reduce the list, we apply the second filtering step.
In this stage, which we introduce in this work, we use Wikipedia to create a link graph: entities as vertices in this graph.
The graph defines the structure of a probabilistic graphical model which we use to rerank the candidate list. We select only top candidates for each mention (2 in our experiments) and still maintain high recall. The two steps are described below.



\subsection{Initial filtering}
\label{subsection dl for el, step 1}

For completeness, we re-describe the filtering technique of Ganea and Hofmann~\shortcite{D17-1276}.
The initial list of candidates is large (see Ganea and Hofmann~\shortcite{P16-1059}, Table 1 for statistics), though there are some mentions (e.g., ``Brexit'' in Figure~\ref{fig intro}) which are not ambiguous. In order to filter this list, besides $p_{wiki}(e|m)$, 
Ganea and Hofmann~\shortcite{D17-1276} use a simple model measuring similarity in the embedding space between an entity and words within the mention span $m$ and a window $c$ around it
\begin{equation*}
q_{wiki}(e | m, c) \propto \exp\lbrace \mathbf{x}^T_e  \sum_{w \in (m, c)} \mathbf{x}_{w} \rbrace,
\end{equation*}
$\mathbf{x}_e$ and $\mathbf{x}_w \in \mathbb{R}^{d_e}$   are external embeddings for entity $e$ and word $w$, respectively.
Note that the word and entity embeddings are not fine-tuned, so the model does not have any free parameters.
They then extract $N_p = 4$ top candidates according to $p_{wiki}(e|m)$ and $N_q = 3$ top candidates according to $q_{wiki}(e | m, c)$ to get the candidate list. For details, we refer to the original paper.
On the development set,  this step yields recall of $97.2\%$. 

\begin{figure}[t!]
\centering
\includegraphics[width=0.49\textwidth]{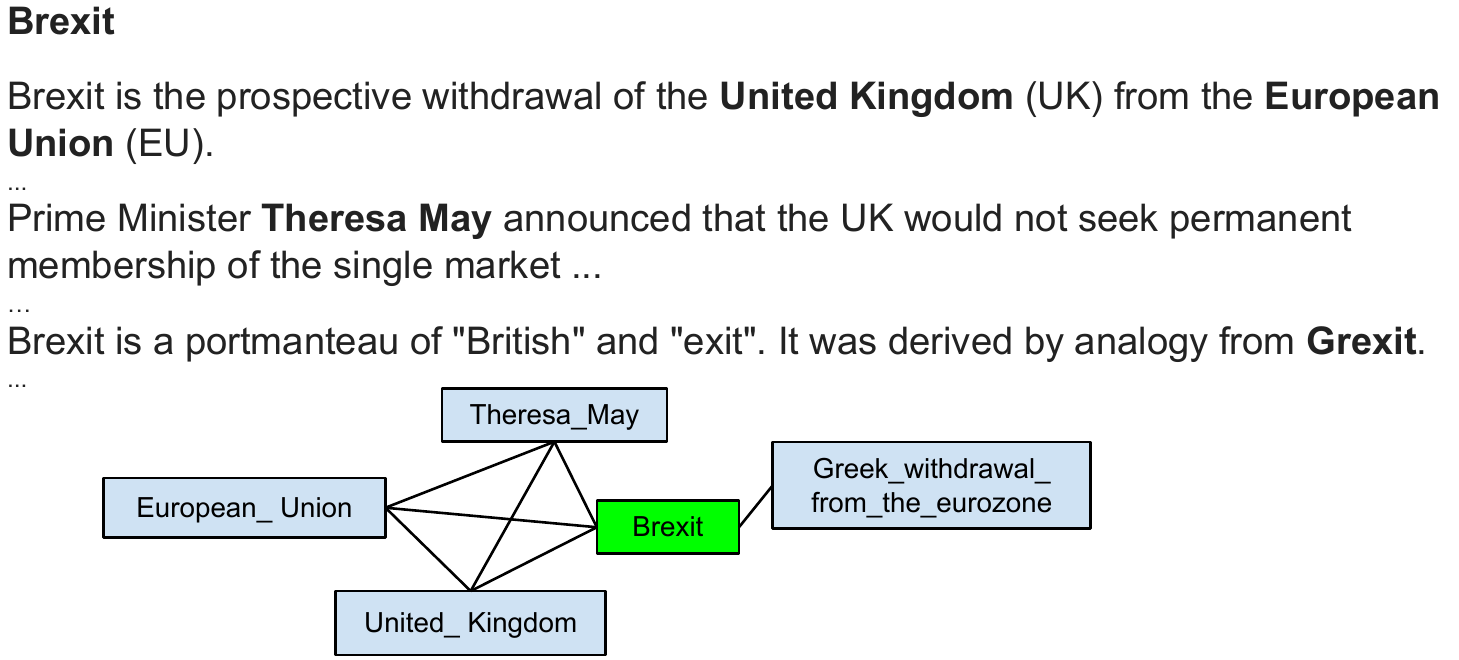}
\caption{A Wikipedia article and the corresponding subgraph of the Wikipedia link graph.}
\label{fig step 2.1}
\end{figure}


\subsection{Message passing on link graph}
\label{subsection dl for el, step 2}
We describe now how we use Wikipedia link statistics to further reduce the candidate list.

\subsubsection{Link graph}
We construct an undirected graph from Wikipedia; 
vertices of this graph are Wikipedia entities. We link 
vertex $e_u$ with vertex $e_v$ if there is a document $D_{wiki}$ in Wikipedia such that either
\begin{itemize}
    \item $D_{wiki}$ is a Wikipedia article describing $e_u$, and $e_v$ appears in it, or
    \item $D_{wiki}$ contains $e_u, e_v$ and there are less than $l$ entities between them.
\end{itemize}
For instance, in Figure~\ref{fig step 2.1}, for document ``Brexit'', 
we link entity \texttt{Brexit} to all other entities. However, we do not link
\texttt{United\_Kingdom} to 
\texttt{Greek\_withdrawal\_from\_the\_eurozone} as they are more than $l$ entities apart.
 
\subsubsection{Model and inference}
\label{subsection step 2.2}
Now we consider unlabeled (non-Wikipedia) documents. We use this step both to preprocess training documents and also apply it to new unlabeled documents at test time.

First, we produce at most $N_q$ + $N_p$ candidates for each mention in a document $D$ as described above.\footnote{Less for entities which are not ambiguous enough.} Then we define a probabilistic model over entities in $D$:
\begin{equation*}
r_{wiki}(e_1, \ldots, e_n|D) \propto \exp \lbrace \sum_{i \neq j} \varphi_{wiki}(e_i, e_j) \rbrace,
\end{equation*}
where $\varphi_{wiki}(e_i, e_j)$ is $0$ if $e_i$ is linked with $e_j$ in the link graph and  $-\Delta$, otherwise ($\Delta \in \mathbb{R}^+$). 
Intuitively, the model scores an assignment $e_1, \ldots, e_n$ according to the number of unlinked pairs in the assignment. We use max-product version of LBP to produce approximate marginals:
\begin{equation*}
r_{wiki}(e_i|D) \approx \max_{\substack{e_1,\ldots,e_{i-1}\\e_{i+1},\ldots,e_n}} r_{wiki}(e_1,\ldots,e_n|D)
\end{equation*} 
For example, in Figure~\ref{fig intro}, we linked \texttt{Donald\_Trump} to \texttt{Brexit} and with \texttt{Theresa\_May}, that are linked in 
the Wikipedia link graph. The assignment {\tt Donald\_Trump}, {\tt Brexit}, {\tt Theresa\_May} does not contain unlinked pairs and will receive the highest score. 

\begin{figure}[t!]
	\centering
	\includegraphics[width=0.45\textwidth]{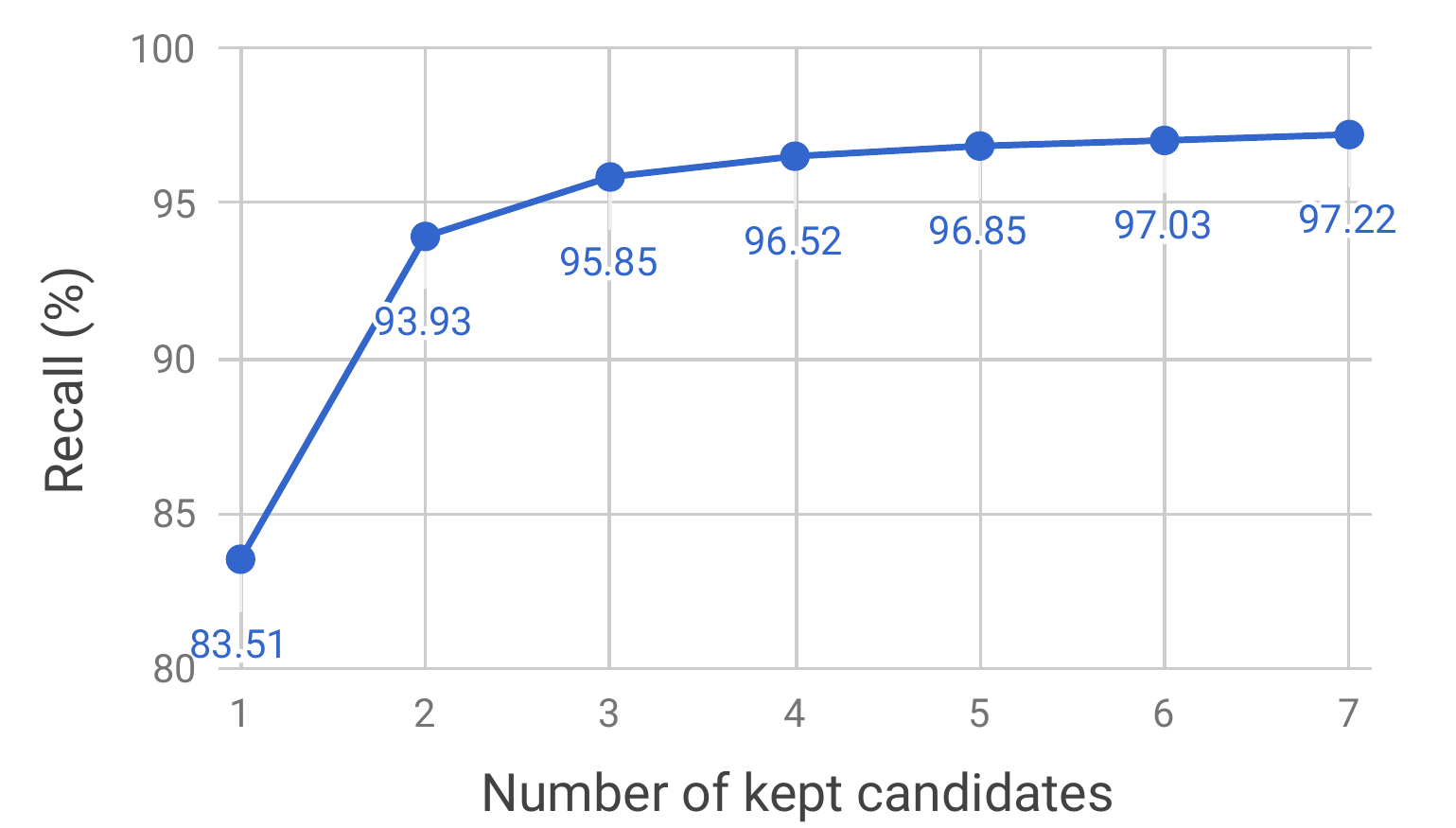}
    \caption{Recall as a function of the candidate number.}
    \label{fig step2 recalls}
\end{figure}

In Figure~\ref{fig step2 recalls}, we plot recall on AIDA CoNLL development set as a function of the candidate number (ranking is according to $r_{wiki}(e_i|D)$). 
We can see that we can reduce $N_p + N_q = 7$ candidates down to $N_w = 2$ and still maintain recall of
$93.9\%$.\footnote{To break ties, 
we chose a mention which is ranked higher in the first step.}
The remaining ($N_p + N_q - N_w$) entities are kept as `negative examples' $E^-_i$ for training the disambiguation model (see Figure~\ref{fig intro}).  

\subsection{Aggregate scoring function}
\label{subsection candidate score}

As we can see from  Figure~\ref{fig step2 recalls}, keeping the top candidate from the list would yield recall of 83.5\%, which is about 10\% below state of the art.    
In order to test how far we can go without using the disambiguation model,
we combine together the signals we relied on in the previous section. Specifically, rather than using 
$r_{wiki}$ alone, we linearly combine the Levenstein edit distance~\cite{1966SPhD...10..707L}, with the scores $p_{wiki}$ and $r_{wiki}$. Parameters are described in the appendix. The coefficients are chosen on the development set. We refer to this score as $s_c(e_i | D)$. 
\section{Experiments}
\label{section experiments}

\subsection{Parameters and Resources}

We used DeepEd\footnote{\url{github.com/dalab/deep-ed}}
from Ganea and Hofmann~\shortcite{D17-1276}
to obtain entity embeddings. 
We also used Word2vec word 
embeddings\footnote{\url{code.google.com/archive/p/word2vec/}} 
to compute the local score function and 
 GloVe embeddings\footnote{\url{nlp.stanford.edu/projects/glove/}} 
within the attention model in Figure~\ref{fig:multi-relation-a}.
Hyper-parameter selection was performed on the AIDA CoNLL development set.
The margin parameters $\delta$ and the learning rate 
were set to 0.1 and $10^{-4}$. 
We use early stopping by halting training when F1 score 
on the development set does not increase after 50,000 updates.
We report the mean and 95\% confidence of the F1 scores using 
five runs of our system. See additional details in the appendix.

The source code and data are publicly available at 
\url{https://github.com/lephong/wnel}.

\subsection{Setting}
We carried out our experiments in the  
standard setting but used other (unlabeled) data for training, as described below. We used six test sets: 
AIDA CoNLL `testb' \cite{hoffart-EtAl:2011:EMNLP} (aka AIDA-B);
MSNBC, AQUAINT, ACE2004, cleaned and updated by Guo and Barbosa~\shortcite{guorobust};
CWEB, WIKI, automatically extracted from Clueweb  \cite{guorobust,gabrilovich2013facc1}.
We use AIDA CoNLL `testa' data (aka AIDA-A) as our development set (216 documents).

In our experiments, we randomly selected 30,000 unlabeled documents from RCV1.
Since we focus on the inductive setting,
we do not include any documents used to create AIDA CoNLL development and test sets in our training set.
In addition, we did not use any articles appearing in WIKI to compute $r_{wiki}.$
We rely on SpaCy\footnote{\url{https://spacy.io/}} 
to extract named entity mentions. 


We compare our model to those systems which were trained on Wikipedia or on 
Wikipedia plus unlabeled documents. They are: Milne and Witten~\shortcite{milne2008learning}, Ratinov et al.~\shortcite{P11-1138}, Hoffart et al.~\shortcite{hoffart-EtAl:2011:EMNLP}, Cheng and Roth~\shortcite{cheng-roth:2013:EMNLP}, Chisholm and Hachey~\shortcite{Q15-1011}, Lazic et al.~\shortcite{TACL637}. 
Note that we are aware of only Lazic et al.~\shortcite{TACL637} which relied on learning from
a combination of Wikipedia and unlabeled documents. 
They use
semi-supervised learning and exploit only local context (i.e. coherence with other entities is not modeled). 

We also compare to recent state-of-the-art systems trained 
supervisedly on Wikipedia and extra supervision or on AIDA CoNLL: Chisholm and Hachey~\shortcite{Q15-1011}, Guo and Barbosa~\shortcite{guorobust}, Globerson et al.~\shortcite{P16-1059}, Yamada et al.~\shortcite{K16-1025}, Ganea and Hofmann~\shortcite{D17-1276}, Le and Titov~\shortcite{2018arXiv180410637L}.
Chisholm and Hachey~\shortcite{Q15-1011} used supervision in the form of links to Wikipedia from non-Wikipedia pages, Wikilinks \cite{singh2012wikilinks}). This annotation can also be regarded as weak or incidental supervision, as it was not created with the entity linking problem in mind.
The others exploited AIDA CoNLL training set.
F1 scores of these systems are taken from
Guo and Barbosa~\shortcite{guorobust}, Ganea and Hofmann~\shortcite{D17-1276} and Le and Titov~\shortcite{2018arXiv180410637L}.

We use the standard metric: `in-knowledge-base' micro F-score, in other words,
F1 of those mentions which can be linked to the  knowledge base.
We report the mean and 95\% confidence of the F1 scores using 
five runs of our system.

\subsection{Results}

The results are shown in Table~\ref{tab:results}. 

First, we compare to systems which relied on Wikipedia and those which used Wikipedia along with unlabeled data (`Wikipedia + unlab'), i.e. the top half of Table~\ref{tab:results}. These methods are comparable to ours, as they use the same type of information as supervision. Our model outperformed all of them on all test sets. One may hypothesize that this is only due to using more powerful feature representations rather than our estimation method or document-level disambiguation. We will address this hypothesis in the ablation studies below.
The approach of Chrisholm and Hachey~\shortcite{TACL637} does not quite fall in this category as, besides information from Wikipedia, they use a large collection of web pages (34 million web links). When evaluated on AIDA-B, their scores are still lower than ours, though significantly higher that those of the previous systems suggesting that web links are indeed valuable. Though we do not exploit web links in our model, in principle, they can be used in the exactly same way as Wikipedia links. We leave it for future work.

Second, we compare to fully-supervised systems, which were estimated on AIDA-CoNLL documents. Recall that every mention in these documents
has been manually annotated or validated by a human expert. 
We distinguish results on a test set taken from AIDA-CoNLL (AIDA-B) and 
the other standard test sets not directly corresponding to the AIDA-CoNLL domain. When tested on the latter, our approach is very effective, on average outperforming fully-supervised techniques.
We would argue that this is the most important set-up and fair to our approach: it is not feasible to obtain labels for every domain of interest and hence, in practice, supervised systems are rarely (if ever) used  in-domain. 
As expected, on the in-domain test set (AIDA-B),  the majority of recent fully-supervised methods are more accurate than our model. However, even on this test set our model is not as far behind, for example, outperforming the system of Guo and Barbosa~\shortcite{guorobust}. 

\begin{table*}[t!]
	\small
    \centering
    \begin{tabular}{c||c||c|c|c|c|c|c}
    	Methods & AIDA-B & MSNBC & AQUAINT & ACE2004 & CWEB & WIKI & Avg \\
        \hline \hline
        \emph{Wikipedia} & & & & & & \\
        \cite{milne2008learning} & - & 78 & 85 & 81 & 64.1 & \textbf{81.7} & 77.96 \\
        \cite{P11-1138} & - & 75 & 83 & 82 & 56.2 & 67.2 & 72.68 \\
        \cite{hoffart-EtAl:2011:EMNLP} & - & 79 & 56 & 80 & 58.6 & 63 & 67.32 \\
        \cite{cheng-roth:2013:EMNLP} & - & 90 & 90 & 86 & 67.5 & 73.4 & 81.38 \\
        \cite{Q15-1011} & 84.9 & - & - & - & - & - & - \\
        \hline
        \emph{Wiki + unlab} & & & & & & & \\
        \cite{TACL637} & 86.4 & - & - & - & - & - \\
        Our model & \textbf{89.66} \textpm 0.16 & \textbf{92.2} \textpm 0.2 & \textbf{90.7} \textpm 0.2 & \textbf{88.1} \textpm 0.0 & \textbf{78.2} \textpm 0.2 & \textbf{81.7} \textpm 0.1 & \textbf{86.18} \\  
        \hline \hline
        \emph{Wiki + Extra supervision} & & & & & & &  \\
        \cite{Q15-1011} & 88.7 & - & - & - & - & - & - \\
        \hline 
        \hline
        \emph{Fully-supervised (Wiki +  } & & & & & & \\
        \emph{AIDA CoNLL train)} & & & & & & \\
        \cite{guorobust} & 89.0 & 92 & 87 & 88 & 77 & \underline{84.5} & 85.7 \\
        \cite{P16-1059} & 91.0 & - & - & - & - & - & - \\
        \cite{K16-1025} & 91.5 & - & - & - & - & - & - \\
        \cite{D17-1276} & 92.22 \textpm 0.14 & 93.7 \textpm 0.1 & 88.5 \textpm 0.4 & 88.5 \textpm 0.3 & 77.9 \textpm 0.1 & 77.5 \textpm 0.1 & 85.22 \\
        \cite{2018arXiv180410637L} & \underline{93.07} \textpm 0.27 & \underline{93.9} \textpm 0.2 & 88.3 \textpm 0.6 & \underline{89.9} \textpm 0.8 & 77.5 \textpm 0.1 & 78.0 \textpm 0.1 & 85.5
  	\end{tabular}
    \caption{F1 scores on six test sets. The last column, Avg, shows the average of F1 scores on 
    MSNBC, AQUAINT, ACE2004, CWEB, and WIKI.}
    \label{tab:results}
\end{table*}

\subsection{Analysis and ablations}

We perform ablations to see contributions of individual modeling decisions, as well as to assess importance of using unlabeled data.



\begin{table}[t!]
    \small
    \centering
    \begin{tabular}{l|c|c|c}
    Our model & AIDA-A & AIDA-B & Avg\\
    \hline 
    weakly-supervised & 88.05 & 89.66 & 86.18 \\
    fully-supervised & & & \\
    $\;\;\;\;$ on Wikipedia & 87.23 & 87.83 & 85.84 \\
    $\;\;\;\;$ on AIDA CoNLL & 91.34 & 91.87 & 84.55  \\
    \end{tabular}
    \caption{F1 scores of our model when it is weakly-supervised and when it 
    is fully-supervised on Wikipedia and on AIDA CoNLL. AIDA-A is our 
    development set. Avg is the average of F1 scores on 
    MSNBC, AQUAINT, ACE2004, CWEB, and WIKI. 
    Each F1 is the mean of five runs.}
    \label{tab:results2}
\end{table}

\paragraph{Is constraint-driven learning effective?}
In this work we advocated for learning our model on unlabeled non-Wikipedia documents and using Wikipedia to constraint the space of potential entity assignments. A simpler alternative would be to learn to directly predict links within Wikipedia documents and ignore unlabeled documents.
Still, in order to show that our learning approach and using unlabeled documents is indeed preferable, we estimate our model on Wikipedia articles. Instead of using the candidate selection step to generate 
list $E^{+}_{i}$, we used the gold entity as singleton $E^{+}_{i}$ in training. 
The results are shown in Table~\ref{tab:results2} (`Wikipedia'). The resulting model is significantly less accurate than the one which used unlabeled documents. The score difference is larger for AIDA-CoNLL test set than for the other 5 test sets. This is not surprising as our unlabeled documents originate from the same domain as AIDA-CoNLL. This  suggests that the scores on the 5 tests could in principle be further improved by incorporating unlabeled documents from the corresponding domains. Additionally we train our model on AIDA-CoNLL, producing its fully-supervised version (`AIDA CoNLL' row in Table~\ref{tab:results2}). Though, as expected, this version is more accurate on AIDA test set, similarly to other fully-supervised methods, it overfits and does not perform that well on the 5 out-of-domain test sets.


\begin{table}[t!]
    \centering
    \begin{tabular}{l|c}
    	Model & AIDA-A \\
        \hline
    	Our model & 88.05 \\
        $\;\;\;\;$ without local & 82.41 \\
    	$\;\;\;\;$ without attention & 86.82 \\
		No disambiguation model ($s_c$) & 86.42 \\	
    \end{tabular}
    \caption{Ablation study on AIDA CoNLL development set. Each F1 score is the mean of five runs.}
    \label{tab:ablation}
\end{table}

As we do not want to test multiple systems on the final test set, we report the remaining ablations on the development set (AIDA-A), 
Table~\ref{tab:ablation}.\footnote{The AIDA CoNLL development set appears harder than the test set, as the numbers 
of all systems tend to be lower \cite{D17-1276,2018arXiv180410637L}.} 

\paragraph{Is the document-level disambiguation model beneficial?}
As described in Section ~\ref{subsection candidate score} (`Aggregate scoring function'), we constructed a  baseline
which only relies on link statistics in Wikipedia as well as string similarity (we refereed to its scoring function as $s_c$).  It appears surprisingly strong, however, we still outperform it by $1.6\%$ (see Table~\ref{tab:ablation}).

\paragraph{Is both local and global disambiguation beneficial?}
When we use only  global coherence (i.e. only second term in expression~(\ref{equ model})) and drop any modeling of local context on the disambiguation stage, 
the performance drops
very substantially (to 82.4\% F1, see Table~\ref{tab:ablation}). This suggests that the local scores are crucial in our model:
an entity should fit its context (e.g., in our running example, `Mrs' is not used to address a Queen). Without using local scores the disambiguation model appears to be even less accurate than our `no-statistical-disambiguation' baseline.  
It is also important to have an accurate global model:
not using global attention results in a 1.2\% drop in performance.

\begin{figure}[t!]
\centering
\includegraphics[width=0.45\textwidth]{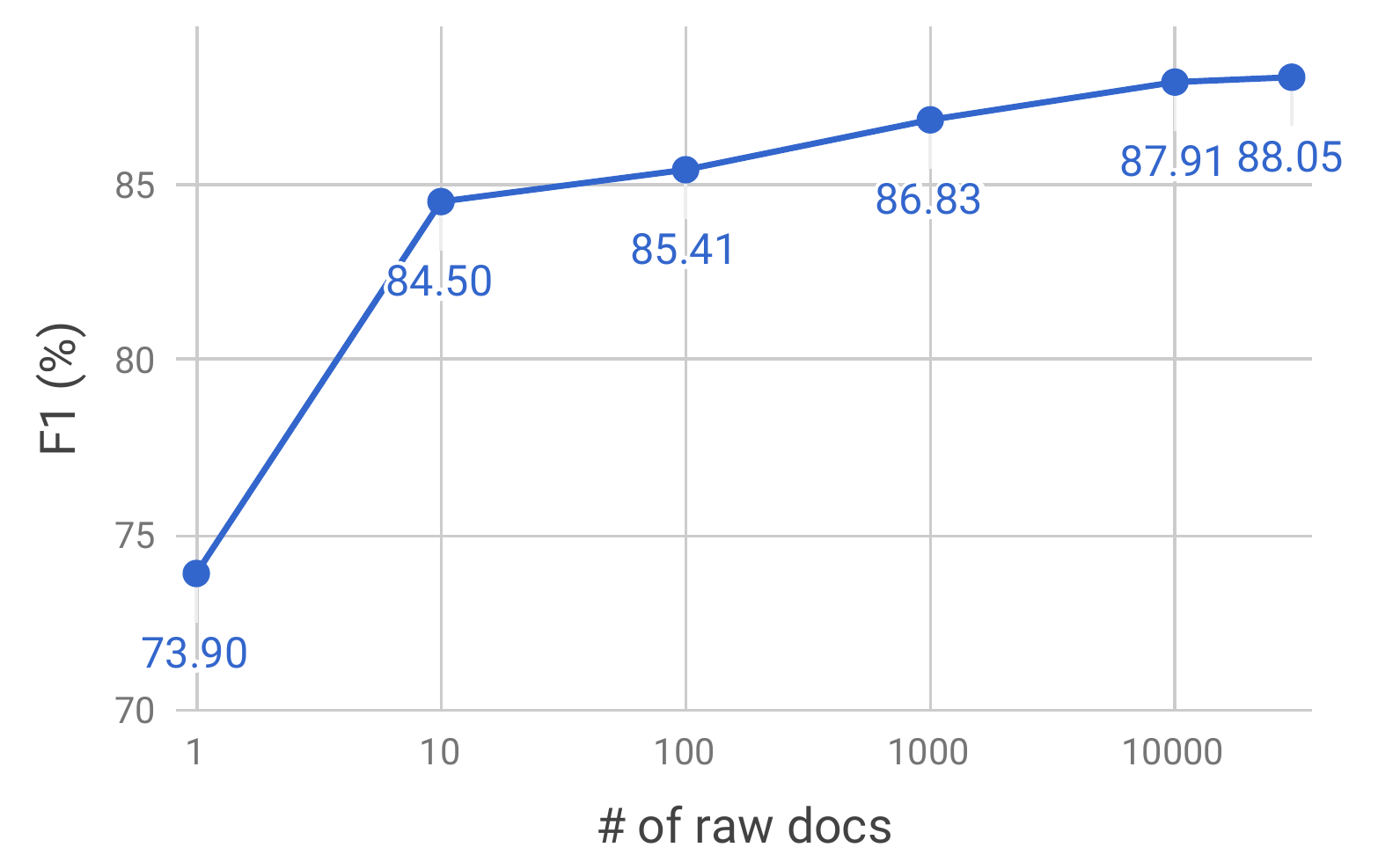}
\vspace{-1ex}
\caption{F1 on AIDA-A vs. 
number of unlabeled documents.}
\label{fig number docs}
\vspace{-1ex}
\end{figure}

\paragraph{Do we need many unlabeled documents?}
Figure~\ref{fig number docs} shows how the F1 score changes when we use 
different numbers of unlabeled documents for training. As expected, the score increases 
with the number of raw documents, but changes very slowly after 10,000
documents. 

\paragraph{Which entities are easier to link?}
Figure~\ref{Tab ner type} shows the accuracy of two systems for different NER
(named entity recognition) types. We consider four types: location (LOC), organization
(ORG), person (PER), and miscellany (MICS). These types are given in CoNLL 2003 dataset, 
which was used as a basis for AIDA CoNLL.\footnote{Note that we do not use 
NER types in our system.}
Our model is accurate for PER, achieving accuracy of about 97\%, only 0.53\% lower than 
the supervised model. However, annotated data appears beneficial for other named-entity types.  One of the harder cases for our model is distinguishing nationalities from languages (e.g., ``English peacemaker'' vs 
``English is spoken in the UK''). Both linking options typically appear in the positive sets simultaneously, so the learning objective  does not encourage the model to distinguish the two. This is one of most frequent mistakes for tag `MISC'.


\begin{table}[t!]
\centering
\begin{tabular}{l|c|c}
Type & Our model & Fully-supervised learning \\
&  & on AIDA CoNLL \\
\hline
LOC & 85.53 & 89.41 \\
MISC & 75.71 & 83.27 \\
ORG & 89.51 & 92.70 \\
PER & 97.20 & 97.73
\end{tabular}
\caption{Accuracy (\%) by NER  type on AIDA-A.}
\label{Tab ner type}
\end{table}

\section{Related work}
\label{section rw}

Using Wikipedia pages to learn linkers (`wikifiers') has been a popular line of research both for named entity linking~\cite{cheng-roth:2013:EMNLP,milne2008learning} and generally entity disambiguation tasks~\cite{ratinov2011local}.
However, since  introduction of the AIDA CoNLL dataset, fully-supervised learning on this dataset became standard for named entity linking,
with supervised systems~\cite{P16-1059,guorobust,K16-1025} outperforming alternatives even on out-of-domain datasets such as MSNBC and ACE2004. Note though that 
supervised systems also  rely on Wikipedia-derived features. 
As an alternative to using Wikipedia pages, links to Wikipedia pages from the general Web were used as  supervision~\cite{singh2012wikilinks}. As far as we are aware, the system of Chisholm and Hachey~\shortcite{Q15-1011} is the only such system evaluated on standard named-entity linking benchmarks, and we compare to them in our experiments. This line of work is potentially complementary to what we propose, as we could use the Web links to construct weak supervision.  

The weakly- or semi-supervised set-up,  which we use, is not  common for entity linking. 
The only other approach which uses a combination of Wikipedia and unlabeled data, as far as we are aware of, is by Lazic et al.~\shortcite{TACL637}. We discussed it and compared to in previous sections.
Our set-up is inspired by distantly-supervised learning in relation extraction \cite{mintz2009distant}. In distant learning, the annotation is automatically (and noisily) induced relying on a knowledge base instead of annotating the data by hand. 
Fan, Zhou, and Zheng~\shortcite{fan2015distant} learned a Freebase linker  using distance supervision.
Their evaluation is non-standard. They also do not attempt to learn a disambiguation model but directly train their system to replicate noisy projected annotations.

Wang et al.~\shortcite{D15-1081} refer to their approach as unsupervised, as they do not use unlabeled data. However,  their method does not involve any learning and relies on matching heuristics. Some aspects of their approach (e.g., using Wikipedia link statitics) resemble our candidate generation stage. So, in principle, their approach could be compared to the `no-disambiguation' baselines ($s_c$) in Table~\ref{tab:ablation}. Their evaluation set-up is not standard.


Our model (but not the estimation method) bears similarities to the approaches of Le and Titov~\shortcite{2018arXiv180410637L} and Globerson at al.~\shortcite{P16-1059}. Both these supervised approaches are global and use attention. 

\section{Conclusions}
In this paper we proposed a weakly-supervised model
for entity linking. The model was trained on unlabeled documents
which were automatically annotated using Wikipedia. 
Our model substantially outperforms previous methods, which used the same form of supervision, and rivals fully-supervised models trained on data specifically annotated for the entity-linking problem. This result may be interpreted as suggesting that human-annotated data is not beneficial for entity linking, given that we have Wikipedia and web links. However, we believe that the two sources of information are likely to be complementary. 

In the future work we would like to consider set-ups where human-annotated data is combined
with naturally occurring one (i.e. distantly-supervised one).  It would also  be interesting to see if mistakes made by fully-supervised systems differ from the ones made by our system and other Wikipedia-based linkers.


\section*{Acknowledgments}
We would like to thank anonymous reviewers for their 
suggestions and comments. The project was supported by the
European Research Council (ERC StG BroadSem
678254), the Dutch National Science Foundation
(NWO VIDI 639.022.518), and an Amazon Web Services
(AWS) grant.

\bibliography{ref}
\bibliographystyle{acl_natbib}

\appendix

\section{Model details}

To compute $\hat{s}$, we combine $s$ with $p_{wiki}$ as below:
\begin{equation}
\label{equ f}
\hat{s}(e_i|D) = f\big( s(e_i|D), p_{wiki}(e_i|m_i) \big)
\end{equation}
where $f$ is a one-hidden layer neural network (in our experiment, the number
of hidden neurons is 100). 

Our final model is the sum of $\hat{s}$ and $s_c$ 
(i.e., $\hat{s} + s_c$)
where $s_c$ is computed by a linear combination of:
\begin{itemize}
\item $d(e_i,m_i)$, the string similarity score between the title of $e_i$ 
and $m_i$, using Levenshtein algorithm,
\item $p_{wiki}(e_i|m_i)$, and 
\item $r_{wiki}(e_i|D)$.
\end{itemize}
In other words we have: 
\begin{align*}
s_{c}(e_i|D) = & \alpha \times d(e_i,m_i) + \\
& \beta \times p_{wiki}(e_i|m_i) + \gamma \times r_{wiki}(e_i|D) \numberthis \label{equ sc}
\end{align*}
We tune $\alpha, \beta, \gamma$ on the development set.

\section{Candidate selection}
In a nutshell, our method to automatically annotate raw texts is summarized
in Algorithm~\ref{alg distant learning}. The algorithm receives a 
list of mentions and contexts 
$D = \{(m_1,c_1), (m_2,c_2),...,(m_M,c_M)\}$.
For each $m_i, c_i$, it will compute a list of positive candidates $E^+_i$ and 
a list of negative candidates $E^-_i$.

\begin{algorithm*}[ht]
	\KwIn{$D = \{(m_1,c_1),...,(m_M,c_M)\}$, $n \in \mathbb{N}$}
 	\KwOut{$(E^+_1, E^-_1), (E^+_2, E^-_2), ..., (E^+_M, E^-_M)$: 
    list of positive and negative candidates}
    \For {$(m_i, c_i) \in D$} {
	    compute $p_{wiki}(e_i|m_i), q_{wiki}(e_i|m_i,c_i)$ and $r_{wiki}(e_i| D)$\;
        
        $E^{30} \leftarrow $ 30 candidates with the highest $p_{wiki}(e_i|m_i)$\;
       
        $E_i \leftarrow$ 4 candidates with the highest $p_{wiki}(e_i|m_i)$
        and 3 candidates with the highest $q_{wiki}(e_i|m_i, c_i)$\
        among $E^{30}$;
        
        $E^+_i \leftarrow$ 2 candidates in $E_i$ 
        with the highest $r_{wiki}(e_i | D)$

        $E^-_i \leftarrow E_i \setminus E^+_i$
    }
	\caption{Automatically annotate a raw document}
    \label{alg distant learning}
\end{algorithm*}

\section{Experiments: hyper-parameter choice}
The values of the model hyper-parameters are shown in Table~\ref{table param}.
For our baseline $s_c$, $\alpha,\beta,\gamma$ are 0.1, 1., and 0.95 respectively. 

\begin{table*}
\centering
\begin{tabular}{l|c}
hyper-parameter & value \\ 
\hline
\emph{Model} & \\ 
$d_e, d_w$ (entity and word embedding dimension) & 300 \\
window size & 50 \\
number of hidden neurons in $f$ (in Equation~\ref{equ f}) & 100 \\
mini-batch size & 1 document \\
$\delta$ (margin) & 0.1 \\
learning rate & 0.001 \\

$\alpha$ (in Equation~\ref{equ sc}) & 0.2  \\
$\beta$ (in Equation~\ref{equ sc}) & 0.2 \\
$\gamma$ (in Equation~\ref{equ sc}) & 0.05 \\
number of updates for early stopping & 50,000 \\
\hline
\emph{Candidate selection} & \\
$l$ (max distance between two entities) & 100\\
$-\Delta$ & -1,000 \\
number of raw document for training & 30,000 \\
$|E^+_i|$ number of kept candidates for training & 2 \\
$|E^+_i|$ number of kept candidates for testing & 3 \\
number of LBP loops & 10
\end{tabular}
\caption{The values of the model hyper-parameters}
\label{table param}
\end{table*}

\end{document}